# COMPARATIVE STUDY ON JUDGMENT TEXT CLASIFICATION FOR TRANSFORMER BASED MODELS

Stanley Kingston S[1], Prassanth A[1], Shrinivas A V S[1], Balamurugan M.S[2][*]

[1,2] School of Electronics Engineering, VIT Chennai, 600127, India;
stanleykingston.s2019@vitstudent.ac.in, prassanth.a2019@vitstudent.ac.in,
shrinivas.avs2019@vitstudent.ac.in

*Email: balamurugan.ms@vit.ac.in

**ABSTRACT:**

This work involves the usage of various NLP models to predict the winner of a particular judgment by the means of text extraction and summarization from a judgment document. These documents are useful when it comes to legal proceedings. One such advantage is that these can be used for citations and precedence reference in Lawsuits and cases which makes a strong argument for their case by the ones using it. When it comes to precedence, it is necessary to refer to an ample number of documents in order to collect legal points with respect to the case. However, reviewing these documents takes a long time to analyze due to the complex word structure and the size of the document. This work involves the comparative study of 6 different self-attention-based transformer models and how they perform when they are being tweaked in 4 different activation functions. These models which are trained with 200 judgement contexts and their results are being judged based on different benchmark parameters. These models finally have a confidence level up to 99% while predicting the judgment. This can be used to get a particular judgment document without spending too much time searching relevant cases and reading them completely.

## [1] INTRODUCTION

On July 11 2022, the supreme court of India delivered a record breaking 44 judgements in a single day. The number might not sound overwhelming considering the 70k pending cases that are yet to be given a verdict. But the fact that it's the supreme judicial authority and the highest court of the Republic of such a populated and diverse country like India, is an achievement in itself. So, this figure is very minimal compared to cases that are happening in state (high court), district and municipal courts which handle lakhs of cases per day across the whole country. Thus, vast and complex is the Indian legal system. This incase implies that the verdicts offered in each and every case varies a lot and the judgment documents are the only source to review them for precedence/reference for related cases.

But certain parts of our system are still largely unorganized particularly when it comes to the judgment documentation system. Even though the judgment documents are digitized and available to a common man for reference, the search methodology is inefficient and the person takes a lot of time to understand the final verdict as the documents don't really follow a specified/standardized format. This basically drains out the curiosity and enthusiasm of using a certain case as precedence in order to win his case. Also, by referring to a judgment doc, a person can gain confidence that he could take down big corporations as well using certain precedents so that he/she may not always require a lawyer for activities of this kind.

So, our work involves text extraction and summarization techniques in order to extract the final judgment and summarize the winner based on the verdict statement which could either be the petitioner or the respondent. By this way, we are making it easier for the particular person to look up the final verdict and just utilize his time to read documents which fulfills his/her purpose of using the judgment document and citing relevant information as precedence in his/her defense in the court of law which might help the person to strengthen their argument.

Text extraction and summarization is extensively used in many modern-day websites and applications in order to deliver/generate precise and concise information from a very large array of related characters/words without essentially losing the meaning/context of the message. The transformers in particular vary from older architecture as it attempts to solve tasks sequentially while also handling its related dependencies.

## [2] SIMILAR WORK

Information extraction is one of the major factors that contributes to fulfil the objective of this proposed system. This section displays a number of journals that have implemented various methods in order to extract selective information from documents of different formats and structures.

An Ontological Framework for Information Extraction (OFIE) model involves extraction of information such as title of the journal, abstract, etc. from research journals. The dataset used here is research journals where they are converted into TXT and XML files for selective extraction of information where TXT extracts text-based information and XML for extracting tabulated information and figures. The information is extracted with the help of F-REGEX which works based on token matching. F-REGEX is used to return approximation results to improve its performance in information extraction. The OFIE model performs the best with precision score 91.2%, recall score 91.6% and F-score 90.1%. In future this work may be extended to automate information extraction through machine learning for different documents. The model can also be improved to extract information in real time. [1]

A patent-based classification of BERT model involves fine tuning the BERT model by selective extraction of information such as title, abstract, etc. to classify them and also determine the best classification index out of International Patent Classification (IPC) and Cooperative Patent Classification (CPC). The dataset used for this particular model is patent documents. PatentBERT model used here is a fine tuned DeepPatent model which is fine tuned to classify based on context. Here PatentBERT outperforms DeepPatent with precision score of 81.75% and also the classification is improved when done with CPC as the index. The model

can be improved by employing different fine-tuning and pre-training methods for future use. [2]

A long text classification of Chinese News involves making use of models such as BERT and CNN for context-based classification. The Chinese Long Text News is provided as input where the context is scaled with the help of Dynamic LEAD (DLn) model and provided to the BERT model for Text-to-Text Encoding. The TTE generated is then provided as input to the Local Feature Convolution Layer (LFC) which extracts Global and Local Features, based on which the context is classified. The model provides an accuracy score of 99%and also shows an improvement of 1.1% in performance relative to the base model. The model can further enhance its performance by implementing sentiment classification. [3]

Technical Support System Assistance involves classification of customer service according to the service templates. Here the user query provided by the customer is used as an input to determine the category of service to be provided to the customer. The user query is pre-processed to remove noisy data and feature extraction is employed to vectorize the input. Models such as TF-IDF, BERT, etc. were used for feature extraction. The vectorized outputs are provided to ML models such as Random Forest, KNN, etc. to classify the query based on service. TF-IDF Vectorization had its best performance in terms of vectorization with an accuracy ranging from 56-69%. After identifying the best model, four more features were added to the dataset for providing a more subjective classification and these features were encoded using one hot encoding. The metadata was then used in the model which provided an increase in accuracy with a score of 72.7% for TF-IDF vectorization using Random Forest. The model can be improved by making use of semi-supervised learning for the unlabeled category. [4]

Test Report Prioritization involves a prioritization of reports for proper study of test results. Here the Test Report documents are provided as input to a Language Technology Platform (LTP) which undergoes several processes to collect keywords and represent the report in Bag of Words (BoW) format. After conversion, a Class of Bug is chosen and the Jaccard Index is calculated with respect to the test report. Here only Single Bug reports are used for similarity classification as they provide more information in comparison to Multi Bug reports. Then the DivClass model is implemented where the risk value (ie) number of keywords used and similarity to the Class of Bug is calculated. If the value is high, they have high priority. It is observed that DivClass outperforms DivRisk when it comes to detecting 75% to 100% of bugs. DivClass is also found to inspect less documents in comparison to other models when it comes to identification of bugs. Tools for prioritization of test reports can be developed from the proposed system for future use. [5]

Aspect Level Sentiment Classification involves classification based on the sentimental polarity of the text. Reviews dataset is used as input as they provide multiple aspects and provide suitable training weights. The pre-trained GloVe helps in vectorization of the words in the context. The proposed system makes use of Aspect Context Interactive Attention (ACIA) to assign weights to the words. Here the weights are provided to each word based on the context they belong to. This means that the same words can have multiple weights based on the aspect of the text. After the weights are determined, the overall weight is calculated and the polarity is assigned. It is found that AOA-MultiACIA performed the best in comparison with the other 13 algorithms. It provides an accuracy score of 82.59% and F1-score of 72.13%. In the case of co-occurrence of aspects, the proposed model was able to handle better in comparison to other models. This model can be improved by use of semantic dependency information to identify the relation between words in a sentence for proper aspect identification. [6]

Legal Named Entity Recognition involves extraction of details such as Petitioner Name, Respondent Name, Court Name, etc. from legal documents. Indian Judgment documents are used as a dataset for this NER model. Sentence Level annotation is performed on the judgment documents. However, to identify the entity rich context, a pre-trained spacy transformer model was used. For training the model, four annotation cycles are performed. The pre-annotation was performed with the help of Roberta+ transition-based parser architecture for each cycle except the first cycle. NER coupled with the Roberta-Base Transformer had the best results with a 92% precision score, a recall of 90.2% and the f1-score resulted in 91.1%. This NLP model may provide a scope to improve on Relationship Extraction in other documents. [7]

Fine-Grained Named Entity Recognition involves assignment of entities based on a large entity set. Due to its extensive annotation set, this study uses the AIDA NER type system created for the DARPA AIDA program. The proposed system involves two stages where coarse grained NER tokenization is performed with the help of BERT where, instance level fine grained classification is used after the classification model. For the Coarse grain model, the multilingual model and English model is used. The two-step multilingual model performs best in comparison to the baseline model with precision score of 88.49%, recall of 90.88% and f1-score of 89.67%. [8]

Few-short Named Entity recognition involves recognition of entities such as name, organization, etc. from a brief context. The SNIPS, Task Oriented Parsing and Google Schema-Guided Dialogue State Tracking datasets were used for training and evaluation. The input is provided to the Proto-Reptile meta-model where the weights are initialized to each and every vector followed by computation of the prototype for each category. In order to get the maximum efficiency from the model, fine tuning is performed where the model is split into parts where 80% is used for prototyping and remaining for back-propagation. To improve upon the challenges presented by Prototypical Networks and Reptile models both the inner and the outer ring have different learning rates. The proposed model performs the best in comparison to other baseline models in terms of f1-scores. The model can be further improved by fine tuning the 'other' category to fit a specific entity effectively. [9]

The authors experimented with their residual BiLSTM model for named entity recognition (NER) on datasets in both Chinese and English to evaluate its performance. They compared the performance of their model to the BiLSTM-CRF model, which is currently the most advanced version for NER. The authors also ran tests to examine the effectiveness of NER with various residual block architectures and residual BiLSTM layer counts. They discovered that both in Chinese and English, their model could significantly enhance NER performance, and that the highest performance was achieved with a 4-layer residual BiLSTM model. The use of a systematic review and meta-analysis approach to find and assess the efficacy of a given approach for treating a certain medical disease is covered in this work. The authors searched multiple databases to identify relevant studies and used strict inclusion and exclusion criteria to select the studies that were eligible for inclusion in the review. They then analyzed the data from the selected studies using statistical methods to determine the overall effect of the intervention. [10]

By employing highly ranked documents to extend the question, query expansion techniques in information retrieval strive to close the linguistic gap between a query and a document, and then using the expanded query to re-rank the search results. However, existing methods are limited in their ability to accurately evaluate the relevance of the information used for expansion, because they do not have powerful models for selecting and re-weighting the expansion information. This can lead to the inclusion of non-relevant information in the expansion, which can pollute the query. The proposed residual BiLSTM model is evaluated on four datasets for named entity recognition: CoNLL-2003 and OntoNotes 5.0 for English and MSRA and Weibo for Chinese. The model takes pre-trained English word vectors and Chinese character vectors as inputs. In some cases, the input is also augmented with BERT for dynamic representation. The model is compared to several baselines and is shown to outperform them on all datasets, achieving F1 of 92.22% on CoNLL-2003 and 89.65% on OntoNotes 5.0. The model also outperforms the residual LSTM model from a previous study on these datasets. When the model is combined with BERT as input, it is shown to be more effective than the baselines on the CoNLL-2003 dataset. [11]

Named entity recognition (NER) is an NLP based task that involves identifying and classifying named entities in text, such as people, locations, and organizations, into predefined categories. In recent years, the deep neural networks (DNNs) have performed well in NER tasks, but they require large amounts of labeled training data to perform well. However, obtaining and manually labeling data can be time-consuming and expensive, particularly for NER tasks that require new labeled datasets depending on the named entity tags. Using a huge amount of unlabeled data and a little amount of labelled data, the authors in this work suggest a way for enhancing NER systems that struggle with a lack of training data. The method involves using the bootstrapping approach to generate labeled data, called "machine-labeled data," through the use of a classifier trained on a constrained amount of labeled data that is then used to predict labels for the unlabeled data. Transfer learning is then used to effectively utilize the machine-labeled data in DNN-based NER models. The method is evaluated on two NER tasks and shows improved performance compared to using a constrained amount of labeled data. [12]

Automated Encoding of Clinical Records involves encoding entire medical records to obtain key information. The dataset used here is clinical documents. Here the dataset is processed with the help of modified MedLEE architecture to map codes for clinical terms. It involves term selection where the particular clinical term is extracted through removal of negated terms and abbreviation expansion. This is followed by term preparation where variants of the terms are added to the encoding table for efficient extraction of related information. Parsing is performed for all the terms to convert the information into structured data which leads to table generation. The final step involves mapping the clinical text based on the codes generated in the table where one or multiple codes can be generated for a single clinical document. The modified MedLEE returned best metric scores in comparison to the baseline model with recall score of 83% for mapping the encoded terms to corresponding UMLS codes and 84% recall for term extraction. The model also had a confidence interval of 95%. The performance of the model can be improved by developing an evaluation design that analyses fine-grained details.[13]

The major methods used in text mining are deep learning and conventional. The main distinction between deep learning and conventional methods is that deep learning learns features from large amounts of data in an automated way, as opposed to using pre-designed features based on prior knowledge, which limits the ability to utilize big data. Deep learning can learn feature representation using millions of parameters, while handcrafted features rely on the knowledge of designers and are not scalable to large datasets. This work discusses mainly text feature extraction methods like Filtering method, Fusion method, Mapping method, Clustering method etc. then talks about different deep learning approach like Autoencoder, Restricted Boltzmann machine, Deep belief network, Convolutional neural network, Recurrent neural network and some other examples.[14]

Ontological Driven Block Summarization involves classification of judgement documents based on similarity. The dataset used is Chinese judgement documents. Two classes of dataset were used for training (ie) crime of traffic accident and crime of dangerous driving. The dataset is then processed to perform domain ontology where the domain knowledge of the document is extracted. This is followed by knowledge block summarization where all the domain ontological terms are stored in the form of knowledge block. The corpus extracted from judgment documents and Chinese texts are used for training where one set is trained to classify based on word2vec vectors and the other set classified based on Word Mover Distance. WMD method is used for calculate the similarity index between two knowledge blocks. The merged ontology based WMD approach returns better evaluation scores in comparison to the traditional approach with a 5.5% improvement in accuracy score. The performance of this model can be further improved by adopting more semantic based approaches for knowledge block summarization.[15]

Bottom-Up Relational Learning of Pattern Matching involves making use of templates and performing pattern matching on sample documents to retrieve required information. The dataset used here is composed of two sources and used for two different applications. The first domain dataset used here involves computer-related job postings data and other domain involves a dataset of seminar announcements. The RAPIER System follows a set of rules in order to extract information and training the model. The information is extracted in the form of pre-filler, slot-filler and post-filler slots. This is done in order to limit the size of the information extracted. In order to set the initial rule list for each slot, learning is done separately with the help of CompressLim to find the acceptable rule. Then rule generalization is performed for random pairs of rules. The generalized rules are then evaluated for its efficiency to extract the required information. This is followed by constraint generalization where the constrains required for rules to function are determined. Based on the rules and constrains, the generalized pattern for extraction is determined. This is followed by specialization phase where the number of pre- and post-fillers to be extracted are attached to the existing generalized rules. The RAPIER model with rule-based extraction outperforms the existing baseline RAPIER models with precision score greater than 80% for job listings template. It also shows consistent performance scores for seminar announcements data. The performance of the model can be improved by modifying the model to handle multi-slot extraction rules.[16]

Bangla-BERT transfer learning involves fine tuning the BERT model to identify Bangla language with the help of limited training dataset. The model is evaluated in terms of the ability to perform NLP tasks such as Sentiment Analysis, Text classification, etc. in comparison to other multilingual BERT models and pre-trained Bangla models. The dataset used here is a collection of long text context in Bangla language retrieved from various internet sources such as journals, news articles, etc. The dataset is then pre-processed to remove foreign language data and noisy information. After structurization the data is split into three sets where the first set remains as raw data to suit the requirement of task, set two is use form pre-training and the final set is helpful for fine tuning. Sub-word tokenization is performed where uncommon words are divided into repeated sub-words to maintain the size of the context and efficient training. This is followed by Byte pair-encoding on the pre-trained set followed by sentence tokenization. The sentence tokens generated are provide to the model for training. Here Bangle-BERT model outperforms the multilingual and pre-trained Bangla models in every NLP task in terms of accuracy with 97.03% for Sentiment Analysis, 99.41% for Text Classification and 99.99% Precision for NER. The model can be improved by fine-tuning to perform specific tasks based on the needs of different domains.[17]

Text processing is an important task that involves working with electronic text. Classifying text documents has been a topic of research for some time now. One approach that has been proposed is a concept-based methodology, which represents the meaning of text instead of individual terms. This reduces the number of features required for classification. The authors of the study employed the Support Vector Machine (SVM) algorithm to classify documents and evaluated the performance of document classification using both original and concept-based features. Their approach using concept-based features led to improved accuracy in document classification. They introduced a feature selection method for text classification, which utilized a concept-based approach to reduce the dimensionality of the feature space with the goal of enhancing classifier performance. The authors compared the performance of this method with that of document classification using the original features and concept-based features, and found that the proposed method led to a significant improvement in classifier performance when compared to the concept-based approach.[18]

In natural language processing, pre-processing is crucial when dealing with text documents to ensure accurate and reliable results. This involves various steps, including tokenization, stemming, and vectorization. Machine learning is often used for text document analysis, with document classification being a typical example. Document classification can be performed either through supervised or unsupervised learning. Categories are predefined and assigned to the respective documents in train data for supervised learning. In contrast, unsupervised learning does not require pre-defined categories, and documents are automatically clustered based on their characteristics. Tokenization is a process of breaking down the given text into individual units called tokens. Tokens can include punctuation marks, words, and numbers. This process helps to determine the frequency of words in the text, which can be used to build models. Additionally, tokens can be tagged by word type for further analysis. Stemming is another pre-processing step that involves finding the root form of a word. This process is often rule-based, and the stem word may not always match the dictionary-based morphological root. There are some common errors associated with stemming, including over-stemming and under-stemming. Over-stemming occurs when words are excessively truncated, which can alter or lose the meaning of the word. Under-stemming happens when two words are stemmed from the same root, even if they have different stems.[19]

Self-attention models like Transformer and BERT have proven to be highly effective in performing tasks in case of natural language processing. These models are often pre-trained on large datasets and then thoroughly fine-tuned to particular tasks that follow afterwards. To assess the performance of their models, the researchers compared them to several different baselines. One baseline they used was BERT-mask, which involved masking of tokens in random fashion for the original sentences in a similar way to self-supervised attention. By comparing their models to this baseline, they could clearly demonstrate how superior were the models of self-supervised attention. Another baseline they used was BERT-EDA, which involved editing training sentences through four different operations: synonym replacement, random insertion, random swap, and random deletion. They also increased the dataset volume to a level comparable to other models and used WordNet as guidance. The authors transformed the aspect-based sentiment analysis task into a sentence-pair classification task by creating a secondary sentence from the aspect. Additionally, they contrasted the effectiveness of single-task and multi-task models. RoBERTa is an improved version of BERT. The researchers noted that the results reported were ensemble models that combined the outputs of several single-task models. In their experiments, they reproduced the results based on the officially released RoBERTa checkpoint.[20]

[3] ROLE OF ARTIFICIAL INTELLIGENCE IN LAW

The law is present in every scramble and pillar in this business world. Every day we come across stuff that has a legal backdrop. From something as simple as a rent agreement to MoU between two different countries. Almost all the virtual business transactions—including sales, purchases, partnerships, mergers, and reorganizations—involve legally binding contracts and its associated rules and regulations. We come across legalities every day that we might not even realize that it exists.

Legal system of every country is vast and complex in its own way. It is estimated that the legal services market is at a valuation close to 1 trillion Dollars globally. So technical advancements in those fields do take time to develop as it is firmly rooted in tradition and notoriously sluggish to adopt new tools and technologies.

Even Though the implementation of AI in the legal system took much time to arrive, it has seen exponential growth in the past few years. The reason being that AI in itself had made substantial growth before being implemented in the legal system. Much of it is due to the fact that the unreliableness that comes with the introduction of technology to any field and the legal system being one of the more critical domains for every government and citizens.

The current rise in AI in general is caused by Moore's Law, a fundamental principle of technology. Based on his observations at Intel, Gordon Moore, a scientist there, issued a forecast in 1965. It was found that since integrated circuits were created, their transistor density had increased by a factor of two every year. According to his law, this pattern will continue, with computer power increasing at a rate of about double every two years while decreasing in price. Simply said, more computers at a lower cost. This provides the foundation for the rapid development of AI capabilities and accessibility, especially when combined with the steadily decreasing cost of storing electronic data.

3.1 Contract Review

Contracts are usually the basic entities of a legal understanding between parties involved. It is considered as the lifeblood of a country's economic system. Mainly business transactions are done via the help of a contract. As this is such an important document, the negotiation and conclusion procedures with the parties involved is a painfully tedious process these days.

The tasks involved in forging a contract are that Lawyers from both sides must painstakingly review, modify, and share redacted papers in what seems like an infinite loop which might easily contain around 1000 pages. This of course is a lengthy process and might hinder the plans of companies if delayed by unforeseen circumstances. This in fact affects their business, revenue and more importantly could break the trust between companies which might result in calling off the undergoing process.

This calls for a massive and yet a complex possibility to fundamentally automate this procedure. Startups such as the [Lawgeex](), [Klarity](), [Clearlaw]() and [LexCheck]() are rigorously striving for excellence in this particular process with a vision of capturing the legal market with the implementation of AI. The companies are creating artificial intelligence (AI) systems that can ingest proposed contracts automatically. These systems use NLP technology to analyze the contracts in their entirety and identify the acceptable and problematic portions of the agreement.

As far as this implementation is concerned, these systems are for now designed to work in conjunction with a human in the mix (i.e., semi-automated) that is, even though it does the analysis, the final decision lies with a human lawyer and he/she makes the final decision and forges the contract document. As NLP technology becomes more sophisticated, it is pretty much possible that this semi-automated task might become fully automated (i.e., the complete procedure being handled by AI from beginning to end) in the near future. These AI programs could be developed to empower, within pre-programmed variables, to forge out legally valid agreements.

Although it might seem like something from the future, big corporations such as Salesforce, Home Depot, and eBay are true and well into this thing already by utilizing contract review services that are powered by AI in their daily operations. It is anticipated that the use of such services will become more widespread in the near future.

3.2 Contract Analysis

The above discussed process of forging, negotiating and signing of a particular contract are merely the initial stages of an agreement. Once a contract has been established between the relevant parties, it might well be a major hassle to adhere to the terms and the conditions that were agreed upon on that particular contract. This challenge is evidently more common amongst organizations of large scale. Numerous internal divisions of these enormous corporations will each have huge loads of open contracts with considerable amounts of distinct counterparties. Managing and reviewing them continually is a gruesome and challenging task. These companies have to hire a lot of folks to do this job which might be costlier on the run,

Once again, AI is being utilized to address this challenge by leveraging natural NLP. These NLP-driven technological solutions are being developed to extract data and place in context, the critical data from a corporation's extensive collection of legal contracts, allowing stakeholders across the organization to easily understand their business commitments. While several fresh startup rivals have entered the market to compete in this space, two well-funded technology businesses building similar platforms are [Kira Systems](Kira Systems) and [Seal Software](Seal Software).

These solutions will open up a wide range of commercial prospects. Sales teams can efficiently track contract renewals, thus increasing company's revenue and the opportunities of upselling the products. The team that handles the procurement can now easily monitor the existing details of signed agreements, giving them the power to renegotiate often and when needed. Regulatory teams can now maintain a complete understanding of a corporate's activities for the reasons related to compliance. Additionally, finance teams can stay prepared for mergers and acquisitions (M&A) and due diligence.

3.3 Litigation Prediction

Litigation is basically a legal procedure of taking a valid dispute to a court of law. Basically, this happens when the concerned parties cannot agree on between themselves between the fair and proper outcome for a dispute, they involved themselves in. It might range from something as small as rental disagreement to as big as the policy disagreement between governments. It is a broad term that describes a long and sometimes complex process.

Currently, only a limited number of AI teams are specifically developing machine learning models that use relevant legal precedents and a case's specific fact pattern as inputs to predict the outcome of pending cases. This is because of the fact that the judgements made are quite complex and some sensitive issues might erupt into some unforeseen complications.

Though there are some certain issues, it is of the expert's opinion that as these predictions become more precise, they are expected to significantly transform the field of law. Companies and law firms are now utilizing these predictions to take proactive measures in planning their litigation strategies, streamlining the settlement negotiation process, and reducing the number of cases that ultimately require a trial. These encourage people associated with AI and law.

With a primary focus on tax law, Toronto-based firm Blue J Legal is one them involved in law by creating an AI-driven prediction engine for legal proceedings. The business claims that its AI can anticipate case outcomes 90% of the time. We might be able to see more such companies blooming in this particular space in a few years as there is an increase in demand for such products by corporates around the world.

3.4 Legal Research

Legal research is another field that is experiencing a growing influence of machine intelligence. Legal research in simple terms is defined as "the process of identifying and retrieving information necessary to support legal decision-making". This in court is usually addressed by the word "precedence". Basically, a lawyer searches for judgements related to the cases he is working on and could cite those judgements in front of the judge in a courtroom in order to strengthen their client's case or appeal. This has been a very effective methodology for a very long time.

Legal research used to be a labor-intensive manual activity, requiring junior firm associates and law students to leaf through various amounts of printed case and judgement documents to identify pertinent precedent. This method went digital in recent years with the introduction of software and personal computing by which these days, lawyers typically conduct research using applications like LexisNexis and Westlaw. Yet these antiquated systems don't have much intelligence beyond basic search capabilities.

**[4] METHODOLOGY**

4.1 Data Collection

Exactly 200 judgment documents are collected from the Indian Kanoon website. Then, we read out each document and understand the context of the judgment. So, we extract the paragraphs from the document where the actual judgment has been given and put it out as a separate column named as context and next to this, the winner of the particular case, either of petitioner or respondent is also mentioned separately in a column named judgment in that csv file. This document is used for training and testing accordingly. We made our dataset open sourced which can be accessed through Kaggle.

4.2 System Setup and working

Google colab is basically used to run the code and the GPU being used is the Tesla T4. The steps involved are as follows: Initially, all the required libraries are downloaded and imported and the data is being mounted on google drive. Then, the particular model is being tokenized by auto tokenization. Then, the keywords i.e., petitioner and respondent are being labeled and the whole context is encoded so that each and every word carries its own numeric value in that context. It is then followed by the crucial step that involves the splitting of the data. The dataset was divided 80%, 20% respectively for training and testing. Then, for each model all the 4 different activation functions are being run for 100 epochs separately and are checked against training, validation losses followed by accuracy and the confidence level of prediction.

4.3 Architecture

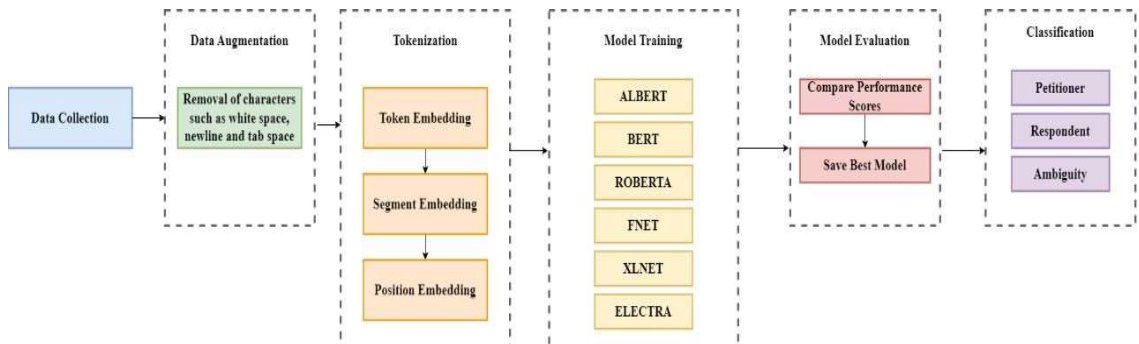

Figure 1: Methodology of the models predicting results

So as mentioned in the above diagram, the data is initially collected and augmented in order to remove unwanted spaces in order to avoid errors. Then the data are being tokenized where in each and every word gets a particular numeric value until the limit if 512 words. These involve 3 various stages of embedding. Then, the dataset is being trained for all the 6 models and are being evaluated by benchmark performance values. The model's best activation function is also being saved. Then, these finally predict the judgement from a doc which is further dived into 3 categories i.e., won by petitioner, won by respondent, ambiguity (neither won by both).

[5] MODELS

5.1 BERT

Machine learning in general had an extensive growth in the past few years. Though each subdomain in ML has had substantial growth, NLP in particular had a steeper growth compared to other subdomains. One particular factor was the introduction of transformers which in turn resulted in the birth of BERT.

Abbreviated as Bidirectional Encoder from Transformer, BERT was initially proposed by google researchers in their work titled "BERT: Pre-training of Deep Bidirectional Transformers for Language Understanding" which was later published in 2019. It is an open-source transformer-based NLP model which specializes in sentiment analysis or natural language understanding. Due to its bidirectional training property of transformers, a wildly used attention model for language modeling, it was proved that it was superior and more advanced than a traditional left-to-right or right-to-left unidirectional training model such as the GPT-3 due to the former's deeper knowledge and understanding of a language's context and flow. This bidirectional training in models was possible with the help of Masked LM model.

BERT comes with two versions of pre-trained models which are namely BERT (base) and BERT (large) both of which are thoroughly trained in an enormous amount of data. BASE is basically used to compare the performance of architecture with the architecture of other models. The LARGE model is capable of producing benchmark results similar to those of the BERT model. Additionally, the LARGE model's utilization of self-supervised learning - where it learns how to identify the underlying language patterns for a given task - allows it to surpass other models in performance. This ability to process language could also be leveraged to improve other models and train them with supervised learning methods.

To further explain the working of the model and transformer in particular, transformer could be best described as an attention mechanism that understands the contextual relation between multiple words in a particular text. The transformer involves the working of two separate mechanisms namely an encoder and a decoder. The former basically intakes and understands the text as input and the latter gives us a prediction as the output for the task. Also, this bidirectional architecture permits the model to grasp and understand the context of the text from every direction, opposed to directional methods which allow only one particular direction.

As regards to the encoder stack, LARGE contained 24 layers whereas BASE only has 12, as was previously stated in respect to the two versions of BERT, LARGE and BASE. These go much beyond the initially disclosed Transformer design (6 encoder layers). In comparison to the original research work on transformer architecture, the LARGE-1024 and BASE-768 feedforward networks and attention heads being 16 and 12 respectively. BERTBASE only has 110M parameters compared to BERTLARGE's 340M. BERT uses two training techniques to define a prediction goal namely MLM (Masked language Model) and NSP (Next Sentence Prediction).

In MLM, the model masks off (hides) 15% of the words in a particular text. The model's job is to try and predict these masked words using the context given by the unmasked words. The output from the encoder must be combined with a classification layer in order to predict the output words. The vocabulary dimension is created by multiplying the embedding matrix by the output vectors, and it is then utilized with SoftMax to calculate the likelihood of each word in the vocabulary. One of the two tasks employed in pre-training BERT, along with the task of predicting the following sentence, is known as masked language modelling.

In NSP, the model is given a particular couple of sentences, and its task is to determine if the second sentence follows the first sentence in the source document. So, while training 50% of the sentences are actual subsequent ones and the other 50% of the sentences are random ones from the doc.

The input goes through a processing step to help the model differentiate between the two sentences before being fed into it for training. A [SEP] token is added for this purpose at the finishing of each sentence, and a [CLS] token is inserted at the beginning of the first sentence. To further identify whether a token fits in Sentence A or Sentence B, each token is given a sentence embedding. The principle of these sentence embeddings is comparable to that of token embeddings, although the vocabulary size is only two. Each token contains a positional embedding that indicates its place in the sequence in addition to the sentence embedding. Positional embedding was first discussed and put into practice in the Transformer work.

Variants
Evolution of this particular open-sourced transformer model led to exploration and advancement backed up with multiple variants. Two such are the ALBERT and RoBERTa.

5.2 ALBERT

Abbreviated as A Light BERT, this model basically has just 12 million parameters as opposed to BERT which has 110 million parameters. So, this makes ALBERT less computationally intensive because of these reduced parameters. These 12 million parameters are present alongside 768 hidden and 128 embedding layers. As one would expect ALBERT was able to reduce the training and inference time. To achieve lesser set of parameters, the following 2 techniques are used:

a. Cross-layer parameter sharing is a technique where only the first encoder's parameter is learned, and then it is shared across all the encoders. These shared parameters can be used for as follows, only feed forward layer or feed forward and multi head attention layer or only multi head attention layer.

b. Factorized embedding layer parameterization: Instead of keeping the embedding layer as many as 768, the embedding layer is reduced to a very minimal amount of 128 layers by factorization.

Besides being light, ALBERT works in the concept of SOP (Sentence Order Prediction) opposed to its parent/predecessor BERT which works on NSP as mentioned earlier. The key difference lies in the fact that while in NSP during training of the model, it receives input pairs of sentences and learns to foretell whether the second sentence will come after the previous one in the original text but in SOP, which is fundamentally a classification model, its goal is to classify whether the two sentences are in the right order or not.

5.3 RoBERTa

Abbreviated as Robustly Optimized BERT pre-training Approach which in many ways is considered superior to BERT itself. The key points of differences are as follows:

a. Dynamic Masking: While RoBERTa mostly employs dynamic masking, BERT primarily employs static masking, in which the same portion of the text is concealed in each epoch. In dynamic masking, certain sentence components are hidden for various Epochs. This approach enhances the model's robustness by exposing it to a variety of masked input configurations during training.

b. Remove NSP Task: It was concluded after many observations and experiments that pre-training the BERT model using the NSP task is considered not particularly useful. When being particularly compared to masked language modelling, it is a much simpler task, which is why it was ineffectual. Both off topic prediction and coherence prediction are combined into a single challenge. Also, because it overlaps with the masked language model loss, the topic prediction component is simple to understand. As a result, NSP will award greater points even in the absence of coherence prediction training. The masked language model also received some noise from the NSP. Therefore, the RoBERTa only works with the MLM task.

c. More data Points: The BERT model is pre-trained on a combination of the "Toronto BookCorpus" and "English Wikipedia datasets," totaling 16 GB of data. On the other hand, RoBERTa is trained on multiple datasets, including CC-News (Common Crawl-News), Open WebText, and others, in addition to the two datasets used in BERT's pre-training. These additional datasets bring the total size to around 160 GB.

d. Large Batch size: RoBERTa utilizes a batch size of 8,000 with 300,000 steps while BERT employs a meagre batch size of 256 with 1 million steps. Which in turn helped to improve multiple factors of the latter model such as the improved speed and performance of RoBERTa compared to that of BERT model.

5.4 XLNet

XLNet is basically a generalized autoregressive pretraining method which works quite similar to BERT but a much better performer compared to its counterpart due to the fact that it essentially captures bi-directional context of a sentence by means of a particular mechanism popularly known as the "permutation language modeling". This model essentially incorporates pretraining concepts from the cutting-edge autoregressive model Transformer-XL. As was previously mentioned, Google uses a model called the transformer for language translation. In essence, it is all about "attention".

The definition of an AR language model is that it is a model that essentially tries to predicts the word to be followed based on the given context or input word. Nonetheless, the input word is restricted to only two directions in this case, which are either forward or backward. The advantages of AR language models include their propensity for success in generative NLP tasks. This is due to the fact that when it develops context, it typically does it either in a forward direction or a backward manner; it cannot do so in both directions simultaneously.

The phases such as the pre-train phase and the fine-tune phase make up any language model. XLNet is primarily concerned with the pre-train stage. During the pre-training phase, a new objective called Permutation Language Modeling was introduced. By building an autoregressive model on all conceivable combinations of the words in a given sentence, PLM aims to capture bidirectional context. XLNET maximizes anticipated log likelihood over all conceivable permutations of the sequence as opposed to fixed left-right or right-left modelling (as in the case of BERT). No [MASK] is required, nor does the input data need to be tainted.

5.5 ELECTRA

Efficiently Learning an Encoder that Accurately Classifies Token Replacements is referred to as ELECTRA. It is a more recent pre-training strategy that basically seeks to match or outperform the downstream performance of its predecessor, a conventional MLM (Masked Learning Model) pre-trained model, while utilizing a lot less compute space resources during the pre-training stage. ELECTRA's pre-training task is based mostly on identifying substituted tokens in the input sequence. Two Transformer types, a generator, and a discriminator are needed for this configuration.

It starts off with replacing some random tokens in a specific input text, the MASK token does this particular job. The generator predicts the original tokens for all masked tokens. The input sequence that was created by swapping out the MASK tokens for the generator predictions is then given to the discriminator. Finally, for each token in the provided input sentence, the discriminator predicts whether or not that particular token is the original one or the one substituted by the generator.

In essence, the main idea is that the generator model is taught to forecast the initial tokens for the MASKed-out ones, whereas the discriminator model is instructed to anticipate which tokens have been replaced in a corrupt sequence.

5.6 FNet

Recent advancements in the field of ML have been possible by the targeting and fiddling of transformer layers. FNet is another one of those models in which the training speed, overall performance has been improved by tweaking the transformer layers. Due to the significant usage of transformers in the field of ML, there have been many advancements with regards to optimizing them. From changing the layers to reducing self-attention layers, to reducing the transformer network sizes greatly. This particular model has been successful due to the fact that it uses a very peculiar way of optimization which involves the use of Fourier transforms.

The Fourier Transform is a statistical concept that can break down a signal into its separate frequencies. However, it does more than just identify the frequencies that are present in the signal. Additionally, it provides each frequency's magnitude in the signal. It is one of the most used transformation methods in signal processing as well as mathematics.

In regards to this topic, FT has been used extensively in:
Solving Partial Differential Equations (using Neural Networks!)
Speeding up convolutions
Stabilizing Recurrent Neural Networks (Fourier RNN and ForeNet)

The most fundamental difference that one might observe is the fact that in the self-attention layer, the usually present encoder in regular architecture is replaced with a Fourier layer that basically applies a 2D Fourier transformation to the input sequence and the hidden dimension. The main intuition behind doing this is the fact that it is considered that the Fourier transform might be able to provide a better way for mixing the input tokens (through its transformation).

From the perspective of time complexity involved, transformers usually have a quadratic complexity ($O(N^2)$) due to the non-linearity of the model and the complexity of self-attention itself. On the other hand, as we know, Fourier transforms are linear and thus offer a much better complexity and an increase in computation speed. Furthermore, the Fourier transform layer doesn't have any learnable weights unlike self-attention, and thus uses less memory because the model size goes down.

5.7 ADAM Optimizer

Optimizers refer to algorithms or methods that are commonly employed to modify the attributes of a deep neural layers, such as weights and learning rate, with the aim of minimizing the associated losses. Each optimizer has their own way of changing the weights and learning rate of the deep neural layers involved for each epoch. There are various types of optimizers that are being used namely the momentum optimizer, adadelta, adagrad, etc… but the one we are using in the work is the 5.7 ADAM optimizer.

Adam, or Adaptive Moment Estimation, is a deep neural network training optimization approach. For the purpose of calculating individual learning rates for various parameters, it mixes first and second order momentums. Adam stores both an exponentially decaying average of the previous squared gradients and an exponentially decaying average of the past gradients, similar to AdaDelta. Since Adam possesses momentum, it can be thought of as a combination of RMSprop and stochastic gradient descent. Adam is a well-liked and very successful optimizer for deep learning models overall.

This model offers several advantages, including its fast convergence rate and the ability to rectify the vanishing learning rate problem, resulting in high variance. The only worrying factor is that it is costly computationally compared to its other counterparts. Nonetheless, Adam is still considered the optimal optimizer for training neural networks in less time with greater efficiency. Although several issues have been raised with the use of Adam in specific situations, academics are still working to find solutions that will allow Adam to produce results that are comparable to SGD with momentum.

5.8 Activation Function

An activation function is a graphical function that determines whether a neuron in a neural network should be activated or not. Its role is critical in predicting outcomes because it decides whether a neuron's contribution to the network is significant or not. Its main task is to produce an output from the input values passed to the node or layer of the network, which allows efficient data flow in the neural network.

Rectified Linear Unit (ReLU)
Function:

$$f(x) = \begin{cases} 0, & \text{if } x \leq 0 \\ x, & \text{if } x > 0 \end{cases}$$

Derivative:

$$f'(x) = \begin{cases} 0, & \text{if } x < 0 \\ 1, & \text{if } x > 0 \\ \text{undefined}, & \text{if } x = 0 \end{cases}$$

Gaussian Error Linear Unit (GELU)
Function:

$$f(x) = \tfrac{1}{2}x \left(1 + erf\left(\tfrac{x}{\sqrt{2}}\right)\right) = x * \Phi(x)$$

Derivative:

$$f'(x) = \Phi(x) + x\phi(x)$$

Sigmoid-Weighted Linear Unit (SiLU) / Swish

Function:

$$f(x) = \frac{x}{1+e^{-x}}$$

Derivative:

$$f'(x) = \frac{1+e^{-x}+x*e^{-x}}{(1+e^{-x})^2}$$

### [6] RESULTS AND OBSERVATION

The performance of every transformer model used is evaluated based on accuracy, training loss and evaluation loss. The activation function with best accuracy is identified for each model and the qualitative scores are identified.

The dataset used for this model comprises three attributes namely the name of the work, the context of the work and the final judgment. Here the dataset was individually fine-tuned to analyze the complex word structures used in the judgment document and identify the final verdict. The dataset contains over 200 documents. The context was encoded for each model and trained. The dataset was split in the ratio of 80:20 for the training and validation set. The training and evaluation were performed for over 100 epochs in case of each model and activation function. Adam Optimization is used and the Attention Dropout is set as 0.3 for all activation functions except Sigmoid-Weighted Linear Unit (SiLU) where the Attention Dropout is set as 0.1.

From table 1 it is observed that models FNET and ROBERTA return better scores over epochs for Rectified Linear Unit (ReLu) activation function. ELECTRA and ALBERT have better scores for Sigmoid Linear Unit (SiLu) activation function. BERT and XLNET have better scores when Gaussian Error Linear Unit (GeLu-New) is used as activation function. The Sigmoid Linear Unit (ReLu) had the worst performance in the majority of the models where it had significantly high training and validation loss of all the hidden activation functions.

Figure 2 and 3 displays the accuracy and training, validation losses of each model respectively. The validation loss increases for all models over epoch. The validation loss decreases until a threshold and starts to increase later. This is because of the lack of sufficient data for fine-tuning where the availability of digital dataset was limited. Other methods to reduce the validation loss were employed but the loss showed no improvement.

From table 1 we can observe that the XLNet-GeLu has the highest accuracy score in comparison to the other models used. However, it is necessary to consider that the performance of the model is better when the training and evaluation loss is low. XLNet-Gelu also has the lowest validation and training loss in comparison to other models as shown in table 1. The XLNet model has the highest computation time in terms of training and evaluation in comparison to other models. The FNet model is the worst performing model in comparison to other models due to higher loss values and poor accuracy score. The performance of the model did not show much improvement with change in activation functions in case of FNet.

From the observations we can arrive at the decision that XLNet is the best performing model in comparison to the other models used based on its accuracy scores and consistency of scores over epochs. XLNet when combined with GeLu activation function can provide the best accuracy score as well as low validation and training loss values.

| Model | Training Loss | Validation Loss | Accuracy |
|---|---|---|---|
| ALBERT-GeLu | 0.982 | 0.844 | 0.65 |
| ALBERT-ReLu | 0.930 | 0.862 | 0.425 |
| **ALBERT-SiLu** | **0.969** | **0.812** | **0.60** |
| ALBERT-GeLu-New | 0.924 | 0.861 | 0.425 |
| BERT-GeLu | 0.590 | 0.611 | 0.75 |
| BERT-ReLu | 1.006 | 0.902 | 0.50 |
| BERT-SiLu | 0.808 | 0.770 | 0.70 |
| **BERT-GeLu-New** | **0.462** | **0.610** | **0.80** |
| ELECTRA-GeLu | 0.843 | 0.808 | 0.725 |
| ELECTRA-ReLu | 0.392 | 0.651 | 0.80 |

| Model | | | |
|---|---|---|---|
| **ELECTRA-SiLu** | **0.555** | **0.579** | **0.80** |
| ELECTRA-GeLu-New | 0.212 | 0.580 | 0.85 |
| FNET-GeLu | 0.495 | 0.730 | 0.70 |
| **FNET-ReLu** | **0.188** | **0.729** | **0.75** |
| FNET-SiLu | 0.815 | 0.862 | 0.625 |
| FNET-GeLu-New | 0.142 | 0.970 | 0.65 |
| ROBERTA-GeLu | 0.727 | 0.652 | 0.70 |
| **ROBERTA-ReLu** | **0.715** | **0.613** | **0.825** |
| ROBERTA-SiLu | 0.581 | 0.713 | 0.775 |
| ROBERTA-GeLu-New | 0.500 | 0.620 | 0.725 |
| **XLNET-GeLu** | **0.177** | **0.416** | **0.85** |
| XLNET-ReLu | 0.656 | 0.630 | 0.80 |
| XLNET-SiLu | 0.681 | 0.652 | 0.75 |
| XLNET-GeLu-New | 0.211 | 0.546 | 0.825 |

Table 1: Benchmark values of all models in each activation function

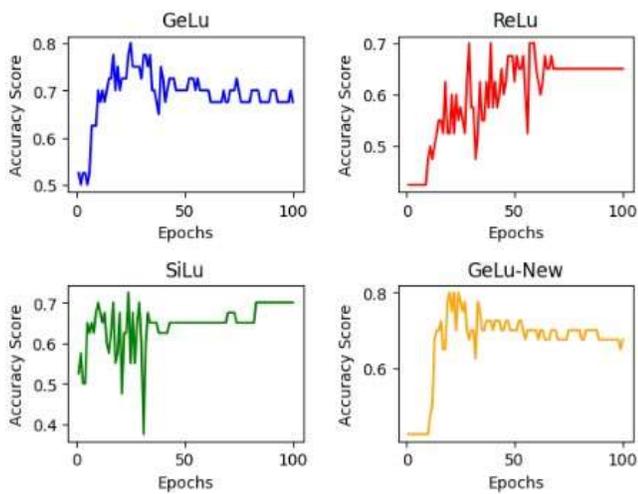
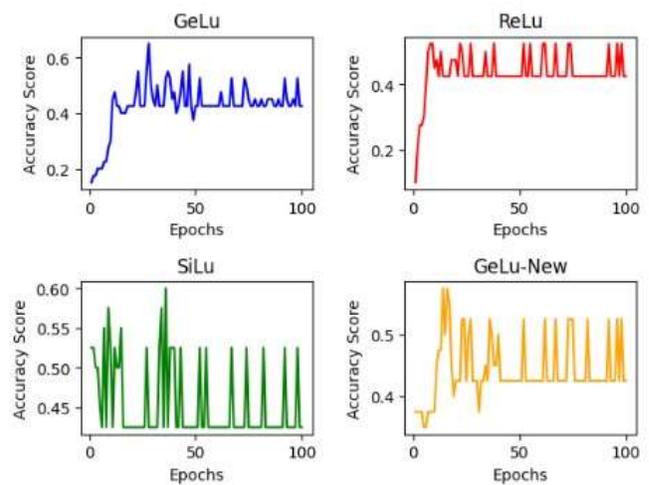
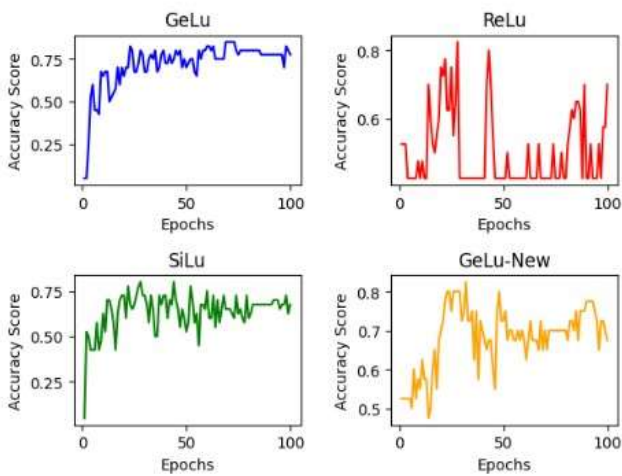
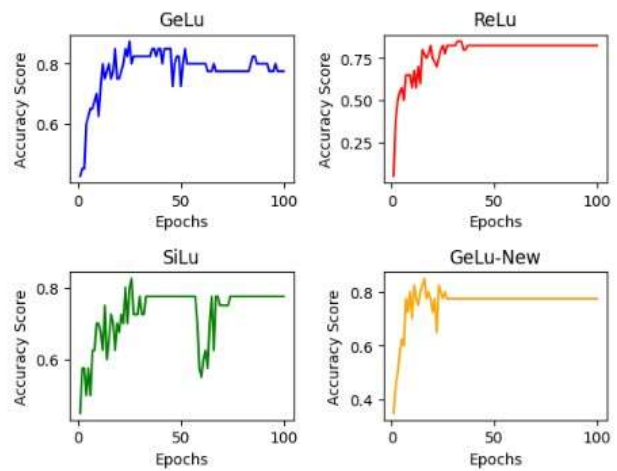
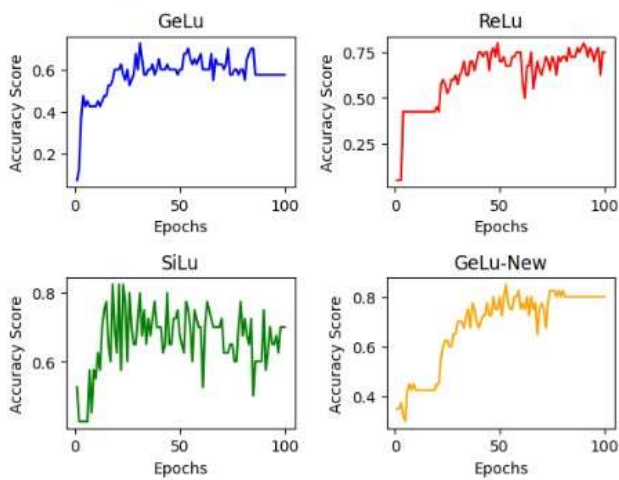
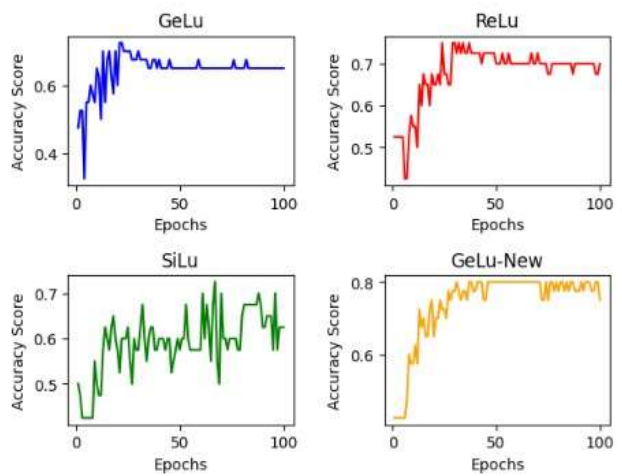

Figure 2-7: Accuracy of the models against each activation functions

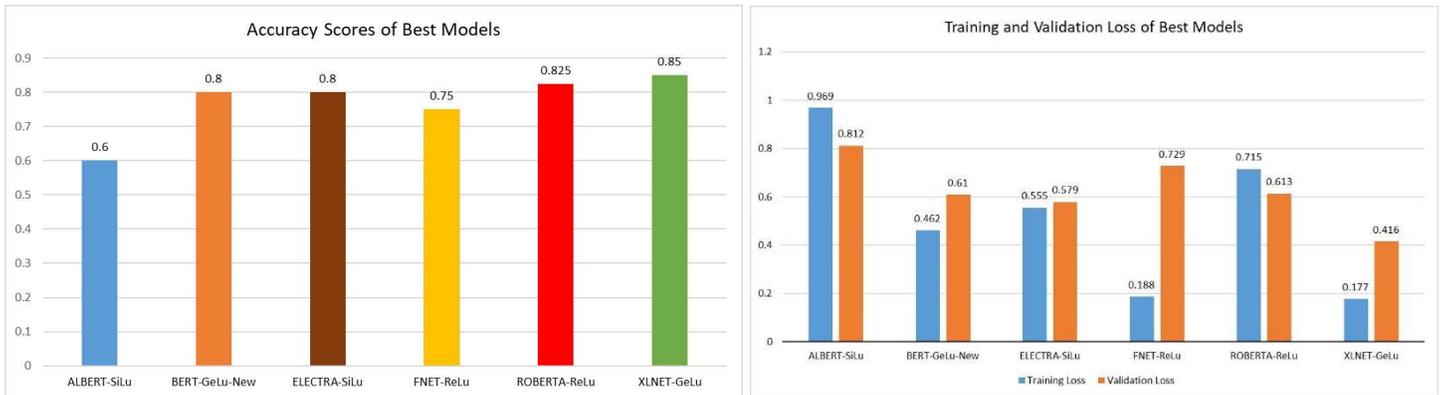

Figure 8-9: Accuracy and losses of best activation function in each model

## [7] CONCLUSION AND FUTURE WORK

This work precisely describes the use of transformers in predicting results from a given context (in this case judgments). Also, the comparative study of different models and activation functions helps us realize how vast and enormous are the layers in various transformer-based ML models. Due to shortage of data, (lesser availability of digitized judgment documents) we weren't able to reduce the validation loss beyond a certain percentage. So, in future with the availability of more data, we could easily reduce the validation loss and bring up a highly optimized model.

Also, this could be extended as a website with search queries and optimization where the user can type out particular keywords which describes the types of parties involved (for ex: bank) and view those particular documents alone for reference and the layers too may try to use it as precedence in a court of law. This work has been uploaded on GitHub.

## [8] REFERENCES